\documentclass[letterpaper, 10 pt, conference]{ieeeconf}  

\IEEEoverridecommandlockouts                              

\newif\ifarxiv
\newif\ifanonymous

\arxivfalse
\anonymousfalse

\arxivtrue      

\usepackage{graphics} 
\usepackage[caption=false,font=footnotesize]{subfig}
\usepackage{epsfig} 
\usepackage{mathptmx} 
\usepackage{times} 
\usepackage{amsmath} 
\usepackage{amssymb}  
\usepackage{cite}
\usepackage[T1]{fontenc}
\usepackage{xurl}
\usepackage{hyperref}
\usepackage{booktabs}
\ifarxiv
\usepackage{tikz}
\fi

\def\eg{\textit{e.g.,\ }}
\def\etc{\textit{etc}}

\newcommand{\projectref}{CRESSim-Neo}
\newcommand{\projectrefbody}{\projectref}
\newcommand{\projectreffirst}{\projectrefbody}
\newcommand{\projecttitleref}{CRESSim-Neo:}
\newcommand{\projectenginesectiontitle}{\projectref\ Simulation Engine}
\ifanonymous
  \renewcommand{\projectref}{The Engine}
  \renewcommand{\projectrefbody}{\textit{\projectref}}
  \renewcommand{\projectreffirst}{\projectrefbody\footnote{Project name omitted for anonymous review.}}
  \renewcommand{\projecttitleref}{}
  \renewcommand{\projectenginesectiontitle}{Simulation Engine}
\fi

\title{\LARGE \bf
\projecttitleref\ A Batched GPU Simulation Engine for Surgical Robotics and Robot Learning
}

\ifanonymous
\author{Anonymous Authors}
\else
\author{{\spaceskip=.9\fontdimen2\font\relax Yafei Ou$^{1}$,\hspace{-0.8pt} Ahnaf Naheen$^{1}$,\hspace{-0.8pt} Tleukhan Mussin$^{1}$,\hspace{-0.8pt} Hans Jarales$^{1}$,\hspace{-0.8pt} Melwin Chacko Moncy$^{2}$,\hspace{-0.8pt} and Mahdi Tavakoli$^{1,2}$}
\thanks{This research was supported by the Canada Foundation for Innovation (CFI), the Natural Sciences and Engineering Research Council (NSERC) of Canada, the Canadian Institutes of Health Research (CIHR), and the Government of Alberta Major Innovation Fund A-MEDICO.}
\thanks{$^{1}$Department of Electrical and Computer Engineering; $^{2}$Department of Biomedical Engineering, University of Alberta, Edmonton, Alberta, Canada. Corresponding author: \texttt{yafei.ou@ualberta.ca}}
\thanks{Code: \href{https://github.com/yafei-ou/CRESSim-Neo}{\nolinkurl{https://github.com/yafei-ou/CRESSim-Neo}}}}
\fi

\begin{document}

\bstctlcite{IEEEexample:BSTcontrol}

\ifarxiv
\AddToHookNext{shipout/foreground}{%
\begin{tikzpicture}[remember picture, overlay]
\node[align=left, xshift=10cm, yshift=-0.8cm] at (current page.north west) {
\begin{minipage}{19cm}
\footnotesize
This work has been submitted to the IEEE for possible publication. Copyright may be transferred without notice, after which this version may no longer be accessible.
\end{minipage}
};
\end{tikzpicture}%
}
\fi

\maketitle


\thispagestyle{empty}
\pagestyle{empty}

\begin{abstract}
We introduce \projectreffirst, a batched GPU simulation engine for surgical robotics and robot learning.
\projectrefbody\ combines position-based simulation of rigid bodies, deformable tissues, fluids, and strands with batched rendering, surgery-specific sensing, and a GPU-resident data pipeline.
The engine supports applications including tissue manipulation, fluid suction, suturing, cable-driven robots, and ultrasound image synthesis.
Direct access to physics and rendering buffers enables GPU-resident robot learning and zero-copy PyTorch integration using DLPack.
We demonstrate \projectrefbody\ across rigid-body, deformable-body, and fluid simulation tasks, including vision-based and surgical robot-learning scenarios.
On an NVIDIA RTX 4090, the engine achieves up to 2.03 million environment steps per second for 8192 parallel CartPole environments, and scales to batched surgical scenarios involving tissue deformation, fluid interaction, and ultrasound sensing.
Overall, \projectrefbody\ provides a unified and scalable platform for surgical simulation, synthetic data generation, and surgical robot learning.

\end{abstract}

\section{Introduction}
\label{sec:introduction}

Recent advances in robotics research have driven growing demand for high-performance, high-fidelity simulators that can serve both as testbeds for robotics control and automation algorithms, and as platforms for generating synthetic training data and experiences.
Despite the availability of numerous robotics simulation platforms, surgical robotics simulation remains particularly challenging because it must capture interactions between various types of objects, including rigid surgical instruments, deformable soft tissues, and fluids.

Typically, a surgical simulator requires one or more physics engines that can solve rigid body, cloth, rod- and thread-like structures, fluids (liquid and gas), and deformable soft body, as well as surgery-specific manipulations such as cutting, cauterization, and fluid suctioning.
Coupling between different simulated objects, such as fluid-rigid body contact, introduces further complexity.
These factors lead to the demand for specific designs and flexibility in both physics and rendering implementations.
Therefore, simulators built on existing physics and/or rendering engines, such as Unity or Isaac Sim, often require additional engineering effort to support missing physics or rendering functionality.
Although this is common practice, integrating new features (\eg adding fluid suctioning support) is usually non-trivial and can sometimes require workarounds that conflict with the engine's default assumptions.

Among physics-based simulation methods, the position-based dynamics (PBD) family of methods \cite{muller2007PositionBasedDynamics,macklin2013PositionBasedFluids,bender2015PositionBasedSimulationMethods,macklin2016XPBDPositionbasedSimulation} is a strong candidate for real-time surgical simulation as it provides a unified and efficient physics framework for diverse materials, including rigid bodies, soft bodies, fluids, and cloth.
Exisiting studies have explored the application of PBD methods to surgical simulation, including cutting \cite{berndt2017EfficientSurgicalCutting,haiderbhai2025Sim2RealRopeCutting}, thread and suturing \cite{xu2018RealtimeInextensibleSurgicala,yu2020RealtimeSuturingSimulation,westergaard2026RealtimeHapticbasedSoft}, soft-tissue deformation \cite{camara2016SoftTissueDeformation,tagliabue2020SoftTissueSimulation}, and fluid irrigation-suction \cite{ou2024LearningAutonomousSurgical}.
iMSTK is an open-source software toolkit for real-time surgical simulations built with PBD \cite{moore2023InteractiveMedicalSimulation}.
In contrast, surgical simulators based on the finite element method (FEM), (\eg SOFA \cite{faure2012SOFAMultiModelFramework,scheikl2023LapGymOpenSource}) and the material point method (MPM) (\eg CRESSim-MPM \cite{ou2025CRESSimMPMMaterial}) are often too computationally expensive for robotics applications.
Nevertheless, despite their advantages, existing PBD-based surgical simulation platforms are mostly designed for surgeon training and typically support only single-scene simulation on the CPU.
Therefore, they lack the scalability and robotics interfaces required for large-scale data generation and learning-based surgical robotics.

\begin{figure*}[t]
    \centering
    \includegraphics[width=1.0\textwidth]{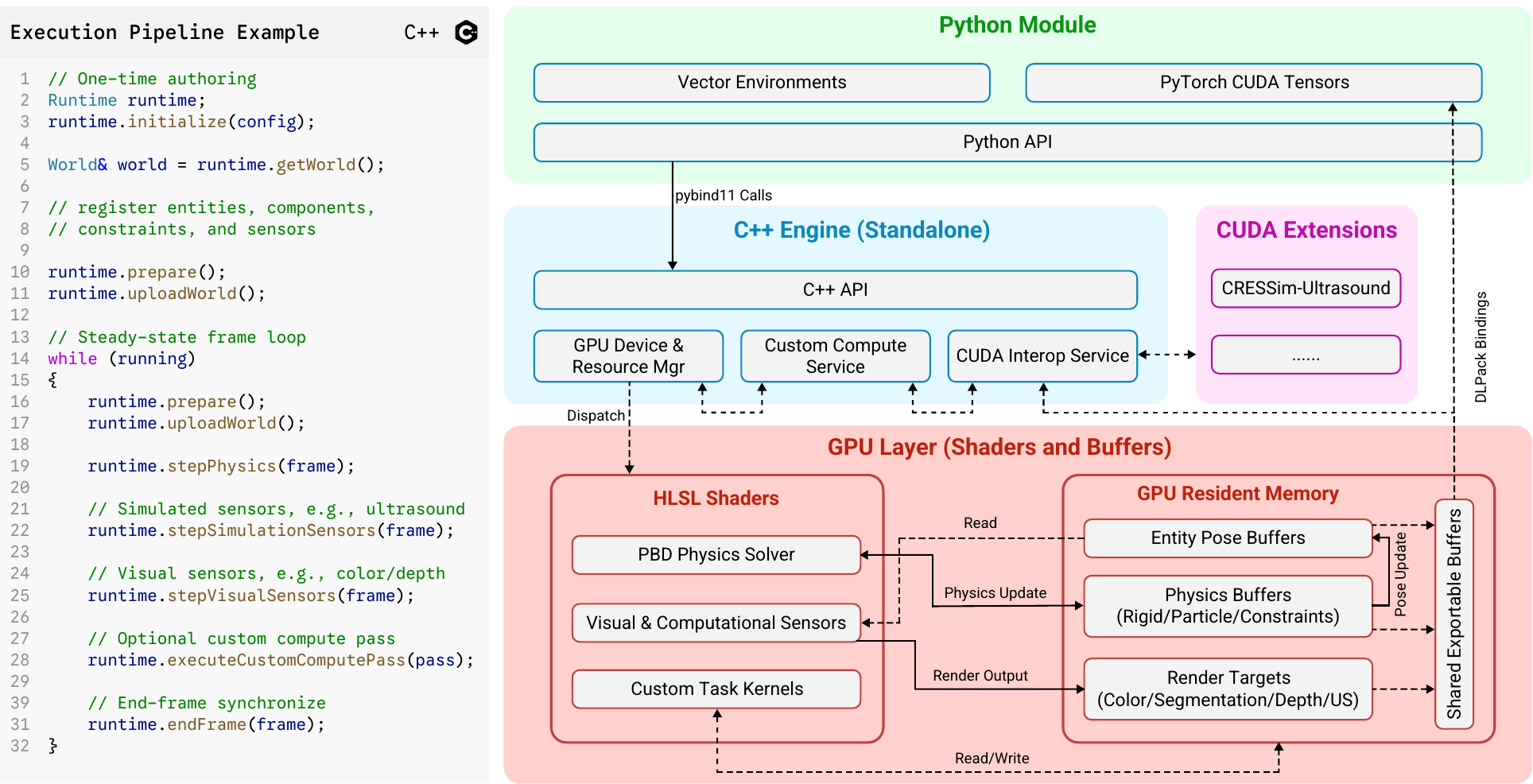}
    \caption{Framework overview of \projectrefbody\ and example code.}
    \label{fig:framework}
\end{figure*}

In parallel, GPU-accelerated robotics simulators have recently emerged to support large-scale robot learning, synthetic data generation, and high-throughput policy evaluation.
Examples such as Isaac Sim, SAPIEN \cite{xiang2020SAPIENSimulAtedPartBased}, MuJoCo Playground \cite{zakka2025MuJoCoPlayground}, Genesis World \cite{genesisaiteam2026RoleSimulationScalable} have demonstrated the potential of scalable simulation for generating high-quality robot locomotion and manipulation data.
They are enabled by GPU-accelerated physics, massively parallel environment execution, GPU-resident learning pipelines, and scalable rendering pipelines, including batched rasterization in some systems and ray-tracing or hybrid renderers in others.

Nevertheless, these systems are generally not designed around the unique physical requirements of surgical scenes.
For example, Isaac Sim and SAPIEN use PhysX, a physics engine focused on rigid-body dynamics with limited soft-body support.
Recent attempts to develop surgical robotics environments on top of such engines therefore often involve trade-offs in physical fidelity or surgical relevance \cite{yu2024OrbitSurgicalOpenSimulationFramework,schmidgall2024SurgicalGymHighperformance}.
Genesis World is a notable exception, offering a unified multiphysics framework written in Taichi Lang.
Isaac Sim is also experimenting with a similar physics backend named Newton,
written in Nvidia Warp\footnote{\href{https://docs.isaacsim.omniverse.nvidia.com/6.0.1/physics/newton_physics.html}{\nolinkurl{https://docs.isaacsim.omniverse.nvidia.com/6.0.1/physics/newton_physics.html}}}.
However, adapting such general-purpose multiphysics capabilities to surgical simulation still requires substantial domain-specific modification, such as for tissue cutting and ultrasound image synthesis.
Thus, existing GPU-accelerated simulators do not yet provide an off-the-shelf solution for surgical robot simulation.

To address this gap, we present \textbf{\projectrefbody}, a surgery-oriented robotics simulation engine that combines the modeling flexibility of surgical simulators with the scalability needed for robot learning and data generation.
The main contributions of this work are as follows:
\begin{itemize}
\item A GPU-accelerated runtime for surgical robotics simulation, supporting rigid bodies, deformable soft bodies, fluids, and strands through a PBD-based physics module and a batched rasterization-based rendering pipeline.
\item Surgery-centric features, including suturing-style interactions, cauterization, cable constraints for continuum and cable-driven robots, and ultrasound image synthesis.
\item A GPU-resident simulation pipeline for high-throughput robot learning and data generation, with customizable compute passes that provide direct access to simulation buffers and CUDA-interoperable outputs.
\end{itemize}




\section{System Overview}
\label{sec:overview}

\subsection{Design Objectives}

The system design is motivated by two core objectives.

\textbf{Multi-Domain Physics and Sensing:}
Surgical workloads require the simulation of rigid instruments, deformable tissues, fluids, and strand-like structures.
Furthermore, the simulator should unify these physical domains with multi-modal observation, including color, depth, semantic segmentation, and simulated ultrasound.

\textbf{GPU-Resident Dataflow:}
Most computational stages should remain GPU-resident.
This includes not only the physics solver and graphics renderer, but also user-defined computations such as observation generation, reward calculation, and post-processing or randomization.
Therefore, interfaces for custom GPU compute, as well as access to internal physics and render-state buffers, must be included.

Additional considerations include (1) cross-operating-system and cross-GPU support, (2) modern graphics API usage, (3) cross-API shading language portability, and (4) graphics and compute API interoperability.

\subsection{System Architecture}

\projectrefbody\ is implemented as a standalone C++ simulation engine with a low-level GPU execution layer and a high-level Python binding layer (Fig.~\ref{fig:framework}).

\textbf{C++ Engine Orchestration:}
This foundational runtime implements the public scene-authoring API, manages the entity-component registry, schedules resource uploads (meshes, textures, sensors), and generates Vulkan compute and graphics commands.
The C++ runtime is fully standalone and can execute independently of any Python bindings.

\textbf{GPU Execution:}
The computational backend is built with Diligent Engine\footnote{\href{https://diligentgraphics.com/diligent-engine/}{\nolinkurl{https://diligentgraphics.com/diligent-engine/}}} and performs all heavy physics and rendering calculations on the GPU.
Vulkan and Direct3D 12 (D3D12) are the two supported graphics API backends.
The PBD solvers, rendering pipelines, and visual/simulation sensors are implemented in HLSL and compiled to compute and graphics shaders.
Custom task logic, such as reward calculation, runs in dispatchable HLSL compute kernels so that performance-critical work also stays on the device.

\textbf{Python Extension:} A lightweight interface layer built on top of the C++ API using \texttt{pybind11}. It exposes the engine runtime API to Python scripts and integrates a Vulkan-CUDA (or D3D12-CUDA, depending on the backend choice) interop layer using DLPack. This allows Python-based learning frameworks, such as PyTorch, to acquire direct, zero-copy access to GPU-resident simulation and sensor buffers.

\subsection{Pipeline Overview}

The engine executes in a staged runtime model with two phases: authoring and stepping.

\textbf{Scene Authoring:} A scene is configured by registering entities, components, and physics constraints through the C++ or Python API.
The engine compiles the required HLSL shaders, allocates device memory, and generates structural layout mapping caches (Section~\ref{sec:resource_exposure}).

\textbf{Frame Stepping:} Once initialized, the runtime steps through a steady-state frame execution loop:
(1) \textbf{data upload} writes dynamic updates (such as kinematic target transforms or user inputs) into device buffers.
(2) \textbf{physics step} dispatches the HLSL PBD physics solvers for rigid-body, deformable constraints, collisions, \etc, on the GPU.
(3) \textbf{sensor and rendering step} processes cameras and simulation sensors (\eg ultrasound synthesis), outputing to GPU render targets.
(4) optional user-defined \textbf{custom compute passes} are dispatched for custom tasks, such as post-processing, data packaging, and post-physics calculations. These passes can be inserted between any stages of the frame loop.
(5) \textbf{frame finalization} synchronizes the GPU and completes any requested data readbacks.

Fig.~\ref{fig:framework} includes an example code block of the full execution pipeline.
This design makes synchronization boundaries explicit and allows custom GPU computations to be interleaved with simulation in a straightforward manner.

\section{\projectenginesectiontitle}
\label{sec:engine}

\subsection{Batched Multi-Environment Scene Representation}

\projectrefbody\ packs multiple environments into one batched scene.
The scene-layout descriptor defines the number of environments and per-environment capacities for renderable objects, lights, and cameras.
An entity is created with an environment index that determines its simulation and rendering state associated with that environment.

On the host side, the scene is authored through a unified entity-component\footnote{We use the term descriptively; the implementation is not an ECS.} world.
Entities may carry transforms, mesh renderers, rigid bodies, colliders, soft bodies, fluids, strands, cameras, lights, or ultrasound-related components.
During \texttt{prepare()} and \texttt{uploadWorld()}, this authored state is converted into GPU-resident scene and physics layouts together with mappings that preserve entity ownership and environment membership.
For renderables, cameras, and lights, the runtime allocates fixed-capacity per-environment slots.
For physics objects, including rigid bodies, colliders, soft bodies, fluids, and strands, the runtime stores environment indices and owner-to-buffer mappings rather than using the same fixed slot scheme as rendering.

\begin{figure}[ht]
    \centering
    \includegraphics[width=1.0\columnwidth]{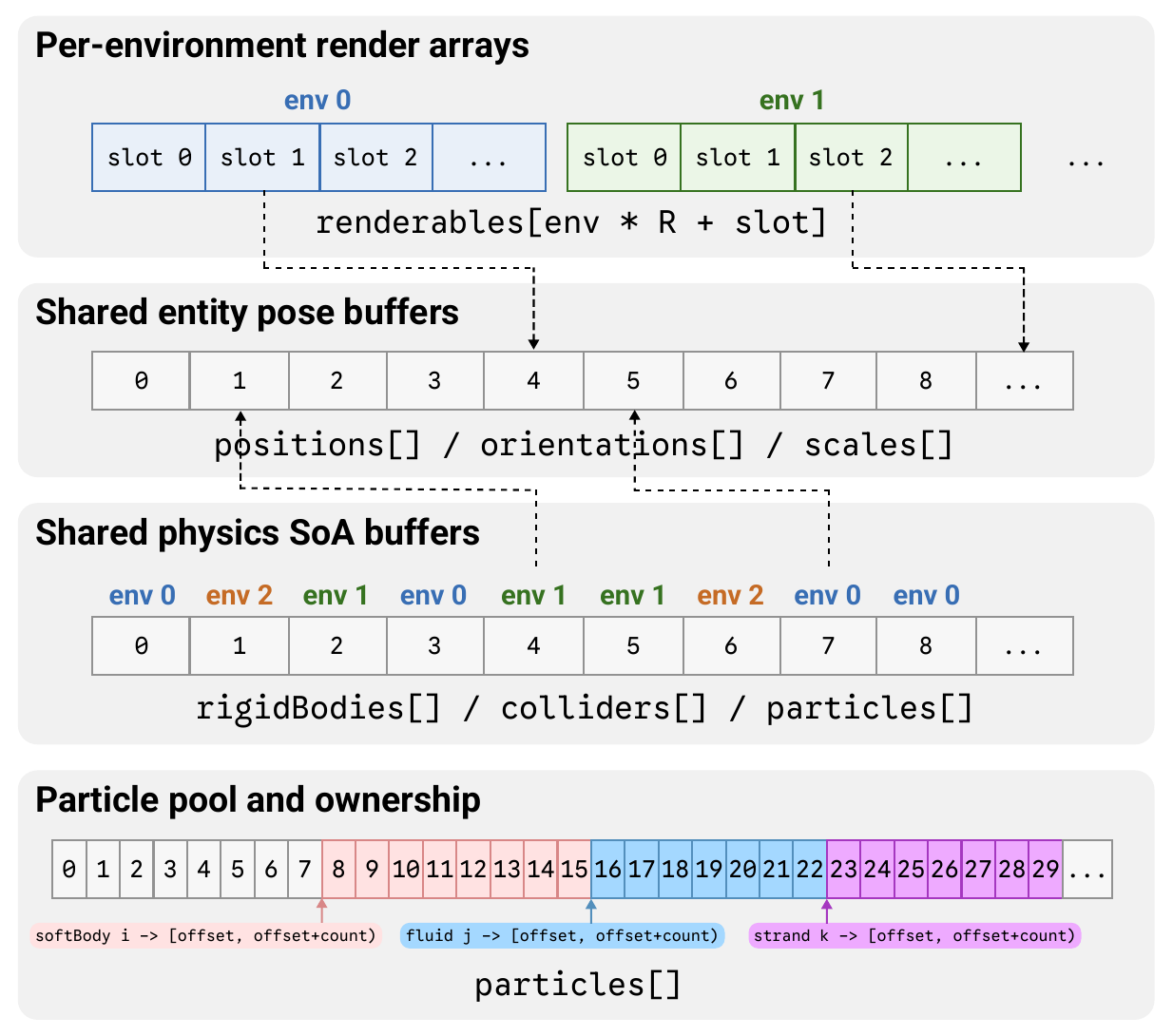}
    \caption{Batched scene data layout.}
    \label{fig:data_layout}
\end{figure}

This representation separates shared resources from per-environment state.
Mesh, texture, material, and shader resources are shared, while dynamic simulation state, per-environment lighting, camera state, environment fluid and IBL settings, and sensor execution are environment-specific.
Fig.~\ref{fig:data_layout} provides an overview of the batched scene data layout.

\subsection{GPU Physics Using PBD}

The physics solver is based on position-based dynamics (PBD), with compliant XPBD-style constraints where needed.
It supports both rigid collider and particle-based dynamics.
Rigid-body simulation supports sphere, box, capsule, and proxy particle colliders, as well as ball, spherical, hinge, and slider joints.
Soft bodies can be represented either by tetrahedral particle models with distance and volumetric constraints, or by meshfree particle models with k-nearest-neighbor distance constraints.
Fluids are handled within the same particle framework using density constraints and viscosity, cohesion, surface tension, and vorticity terms.
The implementations are based largely on the existing CPU-based PBD codebase PositionBasedDynamics\footnote{\href{https://animation.rwth-aachen.de/software/position-based-dynamics/}{\nolinkurl{https://animation.rwth-aachen.de/software/position-based-dynamics/}}}.
Strands are modeled using segment, distance, bend, and twist constraints, enabling cable- and thread-like behavior following the method described in~\cite{liu2023RoboticManipulationDeformable}.

The engine also supports user-authored constraints for surgical task-specific simulations.
Routed cable constraints enforce a path length across guide points attached to rigid bodies, which can be used to simulate cable-driven robots.
Rigid-particle and rigid-strand attachment constraints can connect surgical tools, needles, and threads.
We further support suturing simulation following the method discussed in~\cite{yu2020RealtimeSuturingSimulation}, where strand and rigid proxy particles are coupled with soft tissue through path-following constraints.

\begin{figure}[ht]
    \centering
    \includegraphics[width=1.0\columnwidth]{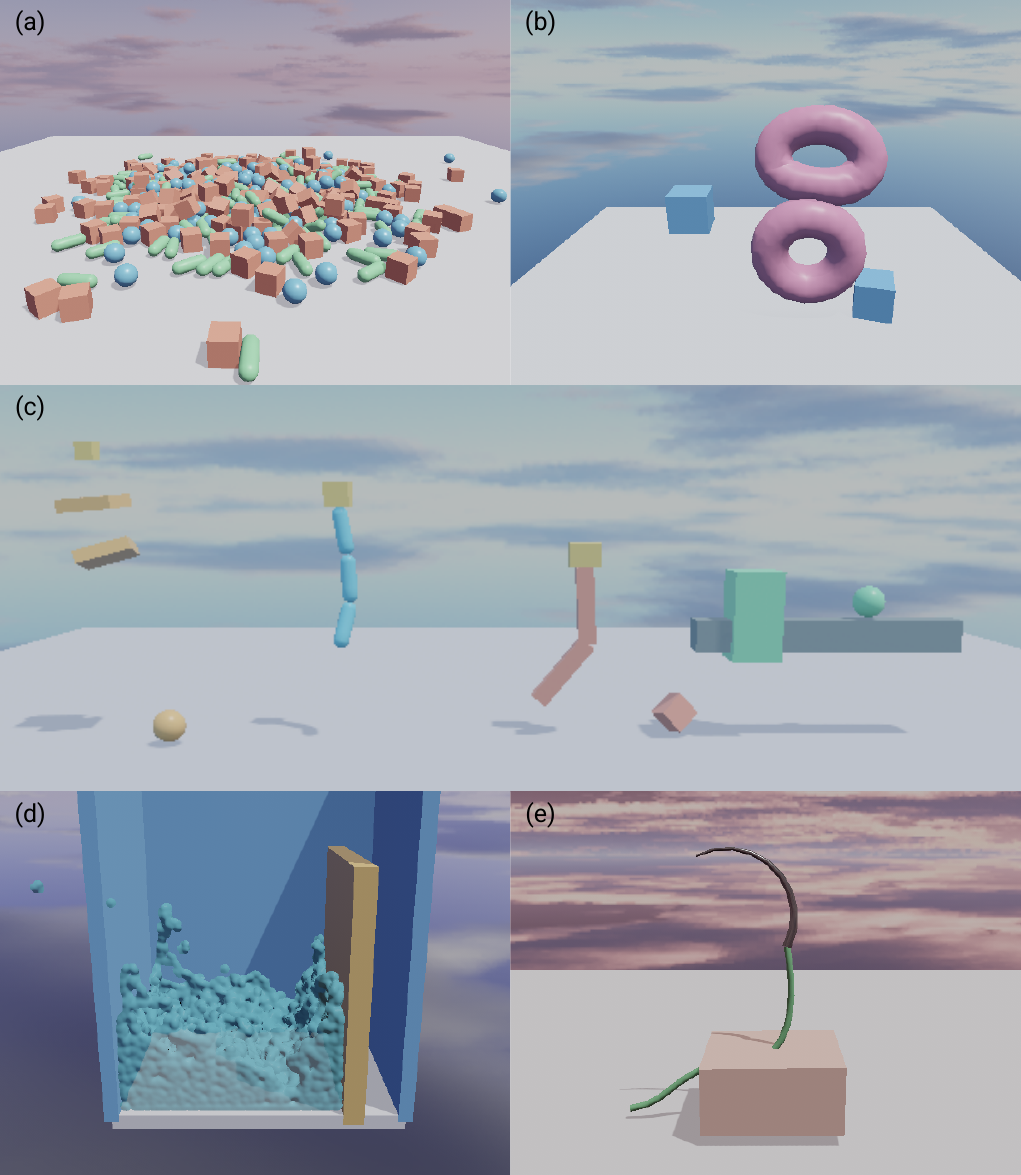}
    \caption{Representative physics scenes: (a) large-scale rigid-body simulation; (b) soft bodies; (c) joints, from left to right: drivable spherical joints, ball joints, hinge joints, and slider joints; (d) fluids; and (e) suturing with a needle, thread, and soft body.}
    \label{fig:physics_demo}
\end{figure}

Because these features share the same parallel prediction, neighborhood and contact generation, and iterative constraint projection pipeline, the physics solver scales well to batched many-environment simulation.
Fig.~\ref{fig:physics_demo} shows example scenes that simulate rigid bodies, soft bodies, fluids, and suturing.

\subsection{GPU Sensing and Rendering}

After authored state is uploaded and physics is stepped, device-side copies are dispatched to update the entity poses based on the physical state.
Visual sensors can then be produced from the current GPU scene state with \texttt{stepVisualSensors()}, and computational sensors such as ultrasound are executed in a separate sensor stage \texttt{stepSimulationSensors()}.

For visual sensing, cameras support three output products: RGB-D, depth-only, and segmentation-with-depth.
Cameras can render either to renderer-managed targets for presentation to screen, or to explicitly defined render targets.
The latter enables cameras with matched configurations to be grouped and rendered in batches into layered array textures spanning multiple environments.
GPU-side camera preparation and indirect drawing further reduce the cost.
Deformable soft-body surfaces and strands are rendered directly from physics-updated GPU buffers, while fluids are composited through dedicated depth, filtering, and color passes, using the method discussed in~\cite{xu2023AnisotropicScreenSpace}.

Ultrasound image synthesis using the COLE algorithm~\cite{storve2017FastSimulationDynamic} is the supported
computational sensor through an external CUDA extension, CRESSim-Ultrasound.
Probe geometry, scanline layout, and image dimensions are defined by the user, and RF data are generated from the current scatterer position.
The scatterers deform together with the underlying soft-body particles,
allowing the ultrasound image to respond in real time to tissue deformation.
The integration of the CUDA extension is discussed in the following section.

\begin{figure}[ht]
    \centering
    \includegraphics[width=1.0\columnwidth]{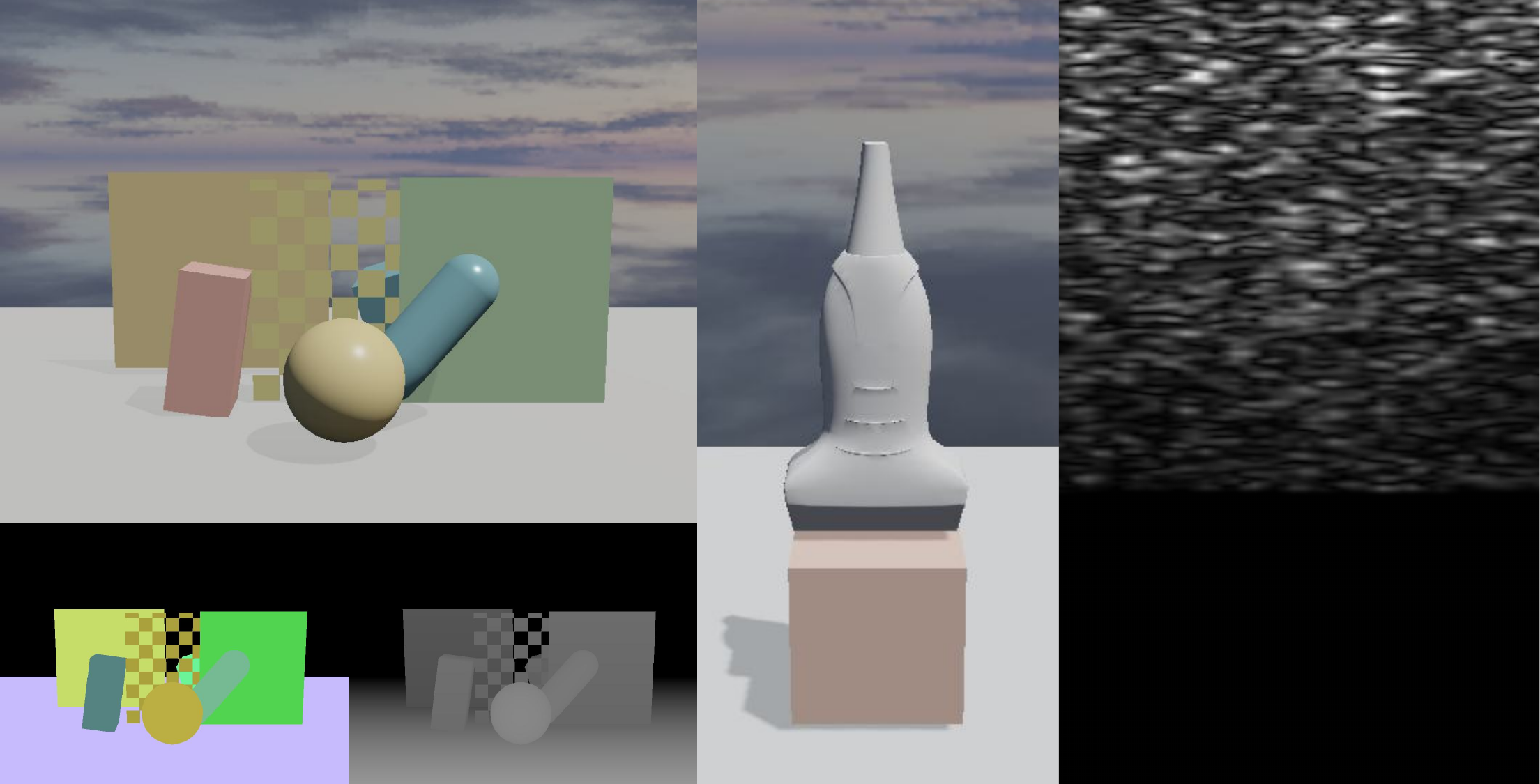}
    \caption{Example sensor outputs. Left: camera RGB rendering with corresponding segmentation and depth products below. Middle: RGB rendering of the ultrasound probe scene. Right: synthesized ultrasound image.}
    \label{fig:graphics_demo}
\end{figure}

Sensor outputs can be consumed either through render-target readback to the host or directly on GPU, enabling downstream custom compute, CUDA/Torch interop, and learning loops without unnecessary host transfers.
Fig.~\ref{fig:graphics_demo} shows example render outputs.

\begin{table*}[ht]
\centering
\caption{Representative GPU resources exposed by the runtime.}
\label{tab:resource_exposure}
\begin{tabular}{ll}
\toprule
Resource & Exposed information \\
\midrule
Prepared rigid layout mapping & Rigid-body and collider indexing \\
Prepared joint layout mapping & Joint indexing and body associations \\
Prepared particle layout mapping & Particle ownership and object ranges \\
Prepared constraint layout mapping & Constraint associations and routed-cable layout \\
Internal simulation buffers & Rigid, joint, particle, and related GPU state \\
Render targets & RGB, depth, segmentation, and ultrasound outputs \\
User-allocated shared buffers & Custom compute inputs, outputs, with DLPack-exportable data \\
\bottomrule
\end{tabular}
\end{table*}

\subsection{Platform Support and CUDA Interoperability}
\label{sec:platform_support_cuda_interop}

The runtime currently targets the Vulkan and D3D12 backends provided by Diligent Engine.
Vulkan is the default cross-platform path and is used on all supported builds, while D3D12 is available on Windows builds.
On macOS, Vulkan support is provided through MoltenVK.

CUDA interoperability can be optionally enabled at build time.
When available, the engine allocates exportable GPU buffers for shared structured data, exposing device pointers for downstream CUDA/PyTorch use.
Synchronization between the graphics/compute backend and CUDA is handled with external timeline semaphores/fences.
If exportable allocation is unavailable, the same APIs fall back to engine-only GPU buffers without interop.

\section{Batched Data Synthesis and Learning Interface}
\label{sec:learning_interface}

\subsection{Resource Exposure and PyTorch Interoperability}
\label{sec:resource_exposure}

Simulation data is exposed through prepared layout mappings, internal GPU resources, and user-allocated shared buffers.
The layout mappings provide the information needed to interpret packed GPU buffers, such as ids, environment indices, per-object particle offsets and counts.

Internal simulation buffers and render targets can be read in custom compute passes.
User-allocated shared buffers can be bound in custom compute passes and, when CUDA interop is enabled, exported through DLPack.
Using the layout mappings, users can interpret simulation state and set up custom compute passes that read internal simulation buffers and render targets, such as RGB outputs, and write task-specific results into user-allocated shared buffers.
Python-side code can then construct \texttt{torch.Tensor} views over those shared buffers.
Table~\ref{tab:resource_exposure} summarizes the main GPU resource categories exposed by the runtime.

\subsection{Custom GPU Task Computation}

Custom GPU task computation is supported through user-defined HLSL compute passes.
These passes are dispatched through \texttt{executeCustomComputePass()} and can be inserted at any point in the staged frame loop, binding internal simulation buffers, shader-readable render targets, and user-allocated shared buffers discussed in Section~\ref{sec:resource_exposure}.
For robot learning environments, this allows actions, observations, rewards, reset masks, and termination flags to be produced and consumed on GPU.

Custom GPU compute is supported independently of CUDA interop (Section~\ref{sec:platform_support_cuda_interop}).
When CUDA interop is disabled, custom compute path remains available, but the user-allocated buffers cannot be exported through DLPack.

\subsection{Example RL Task Instantiations}

\begin{figure}[ht]
    \centering
    \subfloat[\texttt{CartPole}.]{%
        \includegraphics[width=0.42\columnwidth]{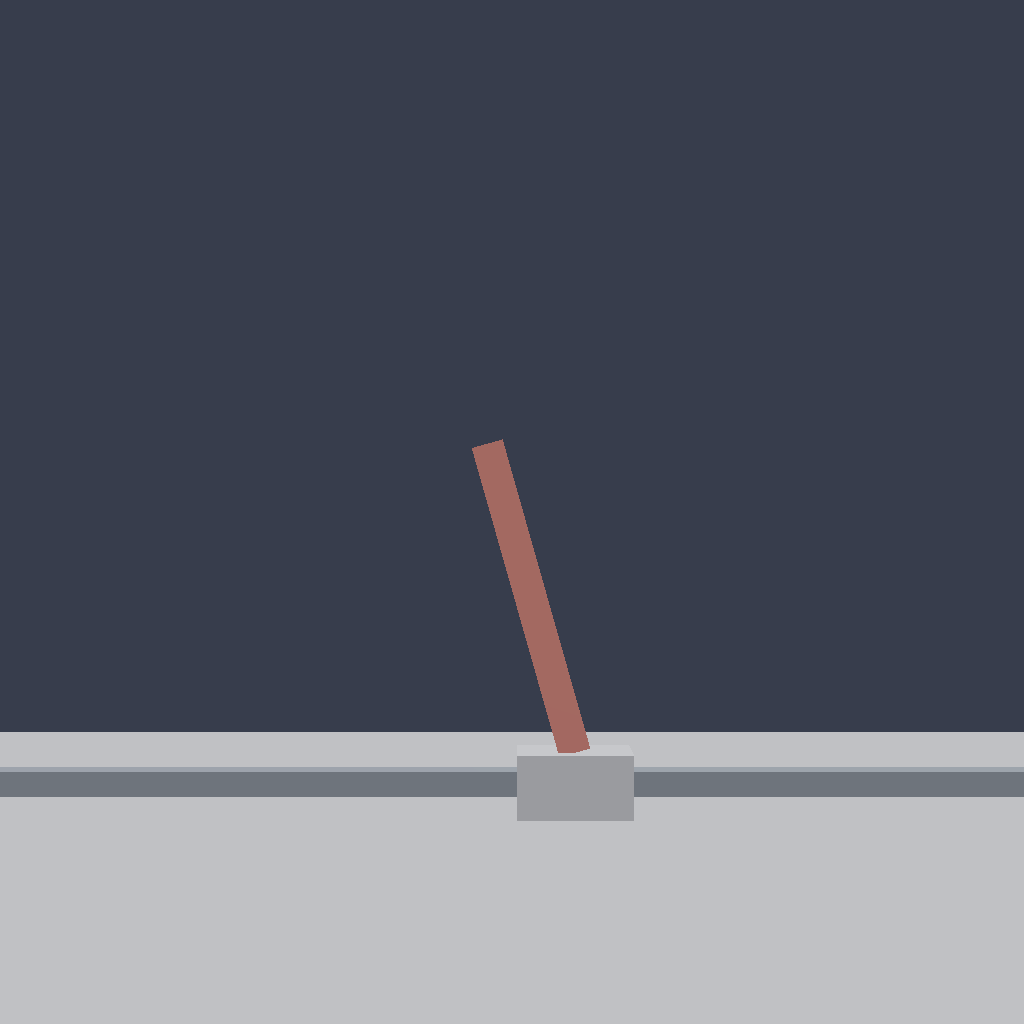}%
    }\hfil
    \subfloat[\texttt{SoftBodyPush}.]{%
        \includegraphics[width=0.42\columnwidth]{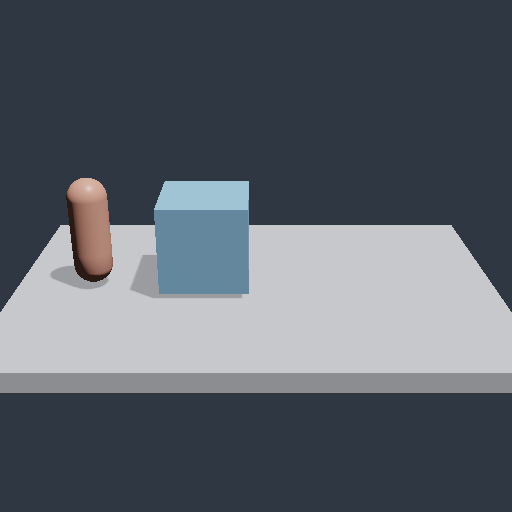}%
    }\\
    \subfloat[\texttt{FluidPour}.]{%
        \includegraphics[width=0.42\columnwidth]{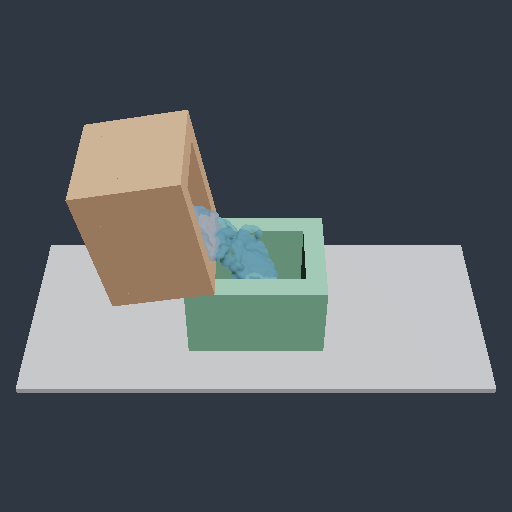}%
    }\hfil
    \subfloat[\texttt{TargetCenter}.]{%
        \includegraphics[width=0.42\columnwidth]{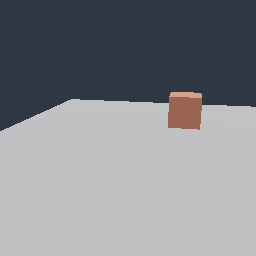}%
    }
    \caption{Representative screenshots of the example RL tasks.}
    \label{fig:example_rl_tasks}
\end{figure}

To showcase the interface described above, we present four simple yet representative batched RL tasks that vary in physics and observation modality: a rigid-body control task, two particle-based manipulation tasks, and an image-based visuomotor task, as shown in Fig.~\ref{fig:example_rl_tasks}.

\textbf{Rigid-body control: \texttt{CartPole}.}
As a minimal rigid-body control example, we implement a batched \texttt{CartPole} task where each environment contains a cart constrained by a slider joint and a pole connected through a hinge joint.
A scalar action specifies the cart actuation target velocity.
Before each physics step, a custom compute pass writes the action into the slider-joint target velocity.
After physics integration, a second pass reads the rigid-body and hinge-joint buffers to construct observations from the cart and pole states and to evaluate reward and termination conditions.

\textbf{Particle-based manipulation: \texttt{SoftBodyPush} and \texttt{FluidPour}.}
To demonstrate tasks that depend on particle state, we introduce two batched manipulation environments where a rigid kinematic actuator interacts with either a soft body or a fluid.
In \texttt{SoftBodyPush}, a 2D action controls the planar motion of a pusher.
After each physics step, the task reads the soft-body particle state to compute observations and rewards, such as the object centroid and how much of the object lies within a target region.
In \texttt{FluidPour}, the action controls the source container's lateral position and tilt.
It then computes observations and rewards from fluid particle statistics, including the fraction of fluid captured by the target container and the fraction spilled outside the workspace.

\textbf{Image-based task: \texttt{TargetCenter}.}
As an example of a visual RL task, we implement a batched \texttt{TargetCenter} environment where the yaw and pitch of a camera is controlled to keep a target near the image center.
A 2-dimensional action updates the camera pose, after which the environment produces RGB outputs into layered render targets.
The observation is the RGB image itself, exported as a GPU tensor.
Reward and termination are computed on GPU from the camera and target poses by evaluating the target's projected displacement from the image center.

\section{Evaluation}
\label{sec:evaluation}

\subsection{Surgical Robotics Capability Demonstrations}

To show the engine's ability to support realistic surgical scenes and robot learning workflows, we present 3 example RL tasks for surgical robotics: \texttt{TissueRetract}, \texttt{BloodSuction}, and \texttt{UltrasoundScan}.

\begin{figure}[ht]
    \centering
    \includegraphics[width=0.85\columnwidth]{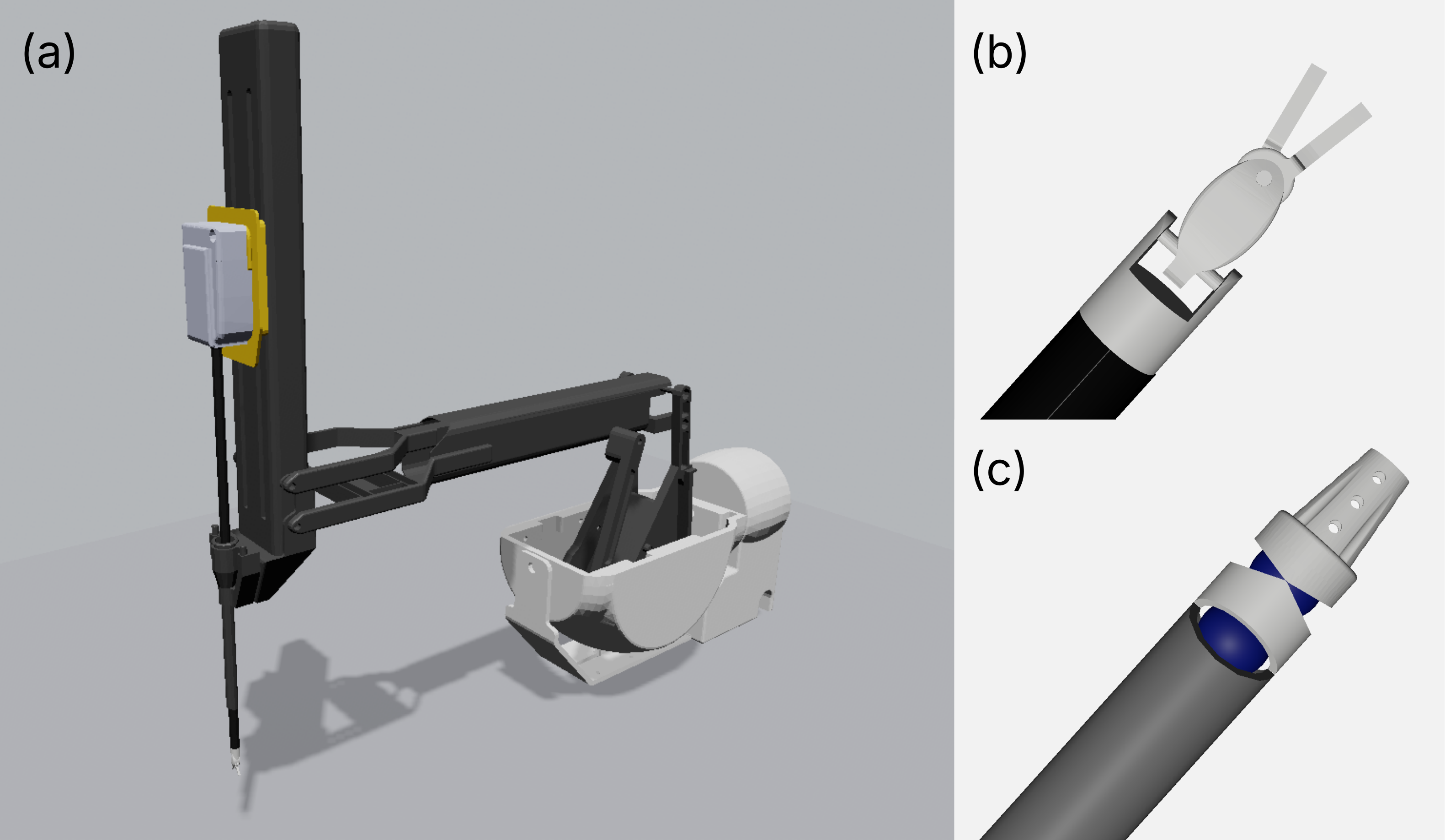}
    \caption{Simulated PSM robot from the dVRK: (a) full robot; (b) Large Needle Driver; (c) Suction/Irrigator.}
    \label{fig:dvrk_psm}
\end{figure}

\begin{figure}[ht]
    \centering
    \includegraphics[width=0.9\columnwidth]{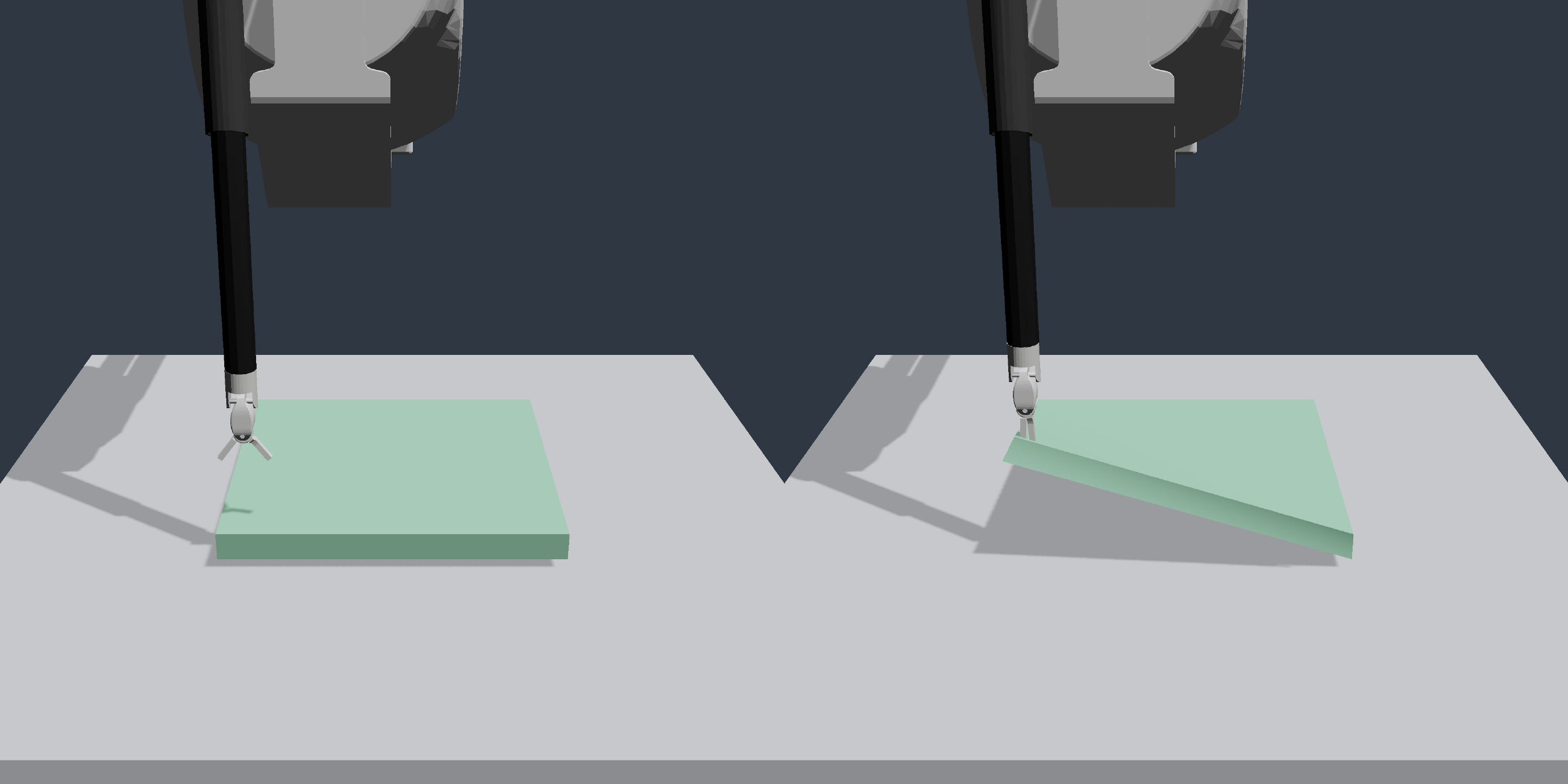}
    \caption{\texttt{TissueRetract}.}
    \label{fig:surgical_task_tissue}
\end{figure}

\begin{figure}[ht]
    \centering
    \includegraphics[width=0.9\columnwidth]{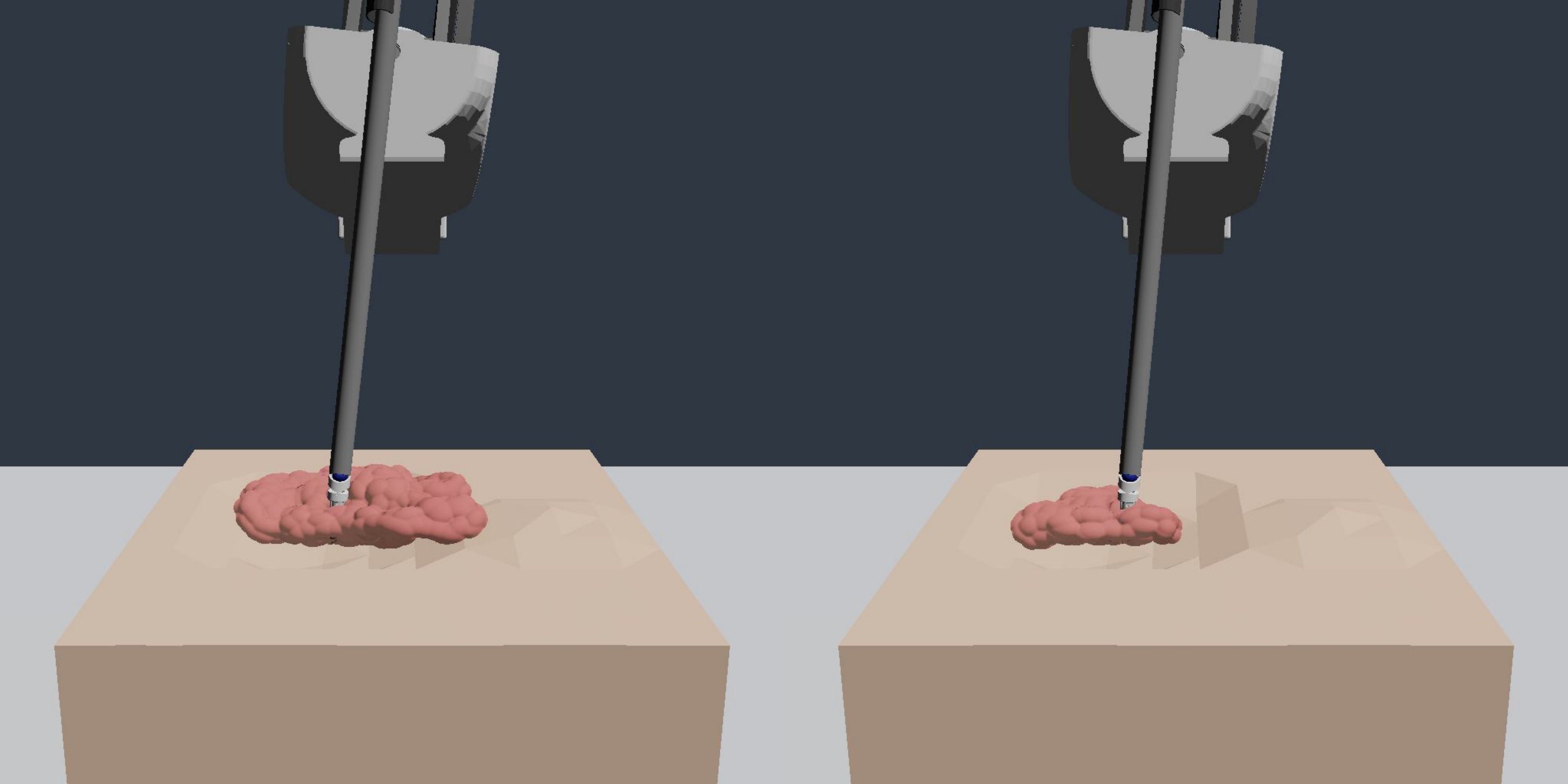}
    \caption{\texttt{BloodSuction}.}
    \label{fig:surgical_task_blood}
\end{figure}

\begin{figure}[ht]
    \centering
    \includegraphics[width=0.9\columnwidth]{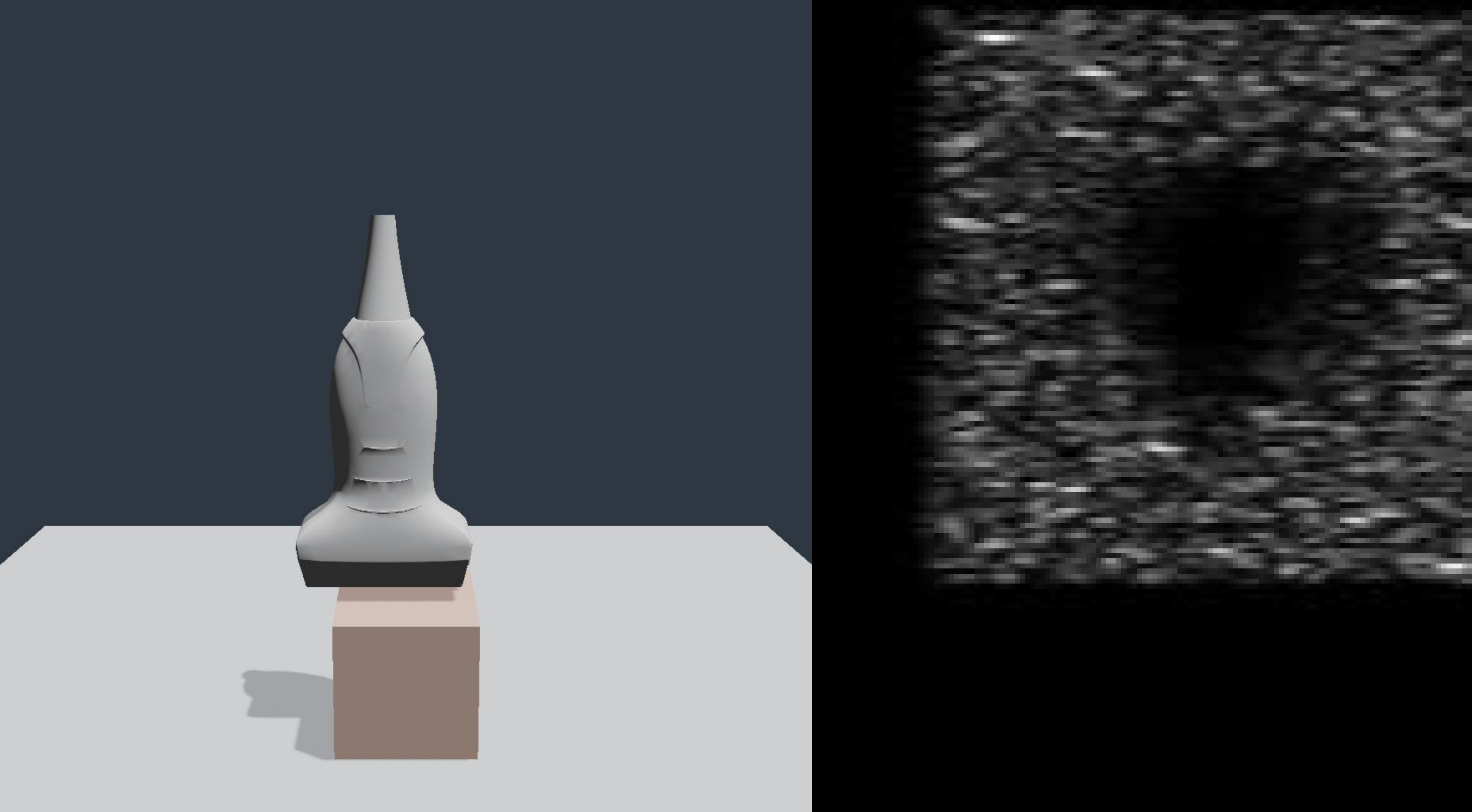}
    \caption{\texttt{UltrasoundScan}.}
    \label{fig:surgical_task_ultrasound}
\end{figure}

With the physics capabilities, we can simulate the patient side manipulator (PSM) from the da Vinci Research Kit~\cite{kazanzides2014OpensourceResearchKit}, as shown in Fig.~\ref{fig:dvrk_psm}.
In the \texttt{TissueRetract} task (Fig.~\ref{fig:surgical_task_tissue}), the PSM tool approaches a selected target location on a deformable tissue patch, establishes a particle-rigid attachment, and lifts the tissue to a target height.
In the \texttt{BloodSuction} task (Fig.~\ref{fig:surgical_task_blood}), a PSM with a suction-irrigator tool interacts with fluid particles inside a deformable container, drawing nearby particles toward the tool and removing particles that reach the suction region.
We further demonstrate a batched \texttt{UltrasoundScan} environment (Fig.~\ref{fig:surgical_task_ultrasound}), where a linear ultrasound probe is moved over a deformable tissue to center a target dark area in the ultrasound image.

\begin{figure}[ht]
    \centering
    \subfloat[]{%
        \includegraphics[width=0.32\columnwidth]{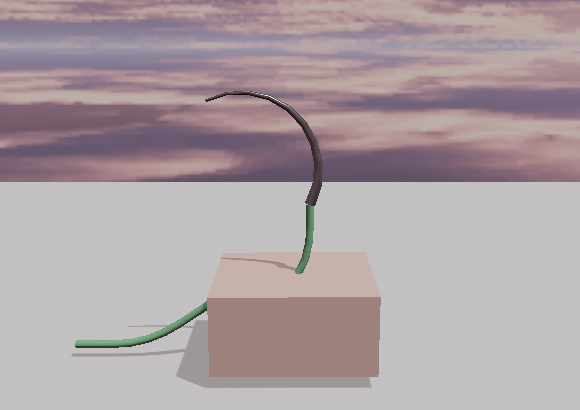}%
    }\hfil
    \subfloat[]{%
        \includegraphics[width=0.32\columnwidth]{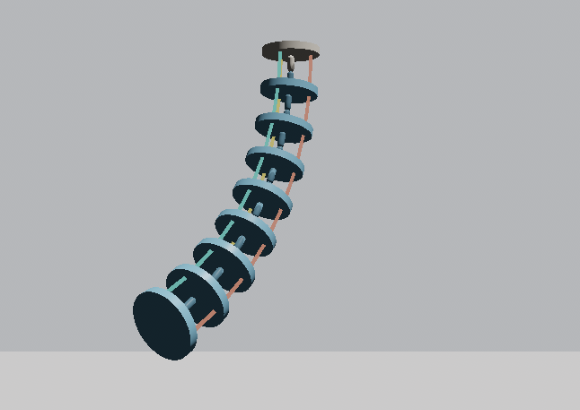}%
    }\hfil
    \subfloat[]{%
        \includegraphics[width=0.32\columnwidth]{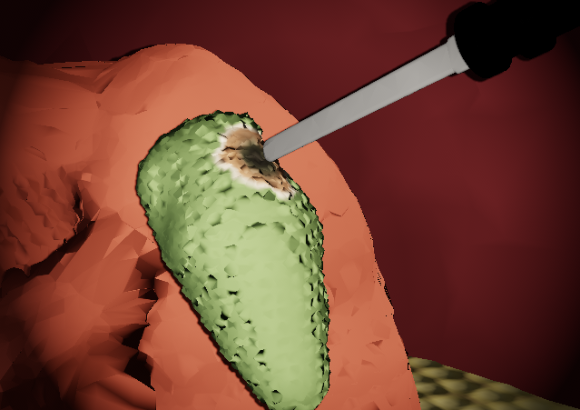}%
    }
    \caption{Additional surgical scenes: (a) soft-body suturing; (b) an approximated CDCR with spherical joints modeling the backbone; and (c) thermal coagulation with varying tissue properties and appearance under heat transfer.}
    \label{fig:surgical_task_additional}
\end{figure}

Beyond these RL environments, we also present additional surgery-relevant scenes, including suturing with a curved needle and thread passing through deformable tissue, cable-driven continuum robot (CDCR) demonstrations based on routed cable constraints, and thermal coagulation with varying tissue properties under heat transfer, as shown in Fig.~\ref{fig:surgical_task_additional}.

\subsection{Simulation and RL Throughput}

\begin{figure*}[t]
\centering
\subfloat[\label{fig:throughput_simulation}]{
    \includegraphics[width=0.48\textwidth]{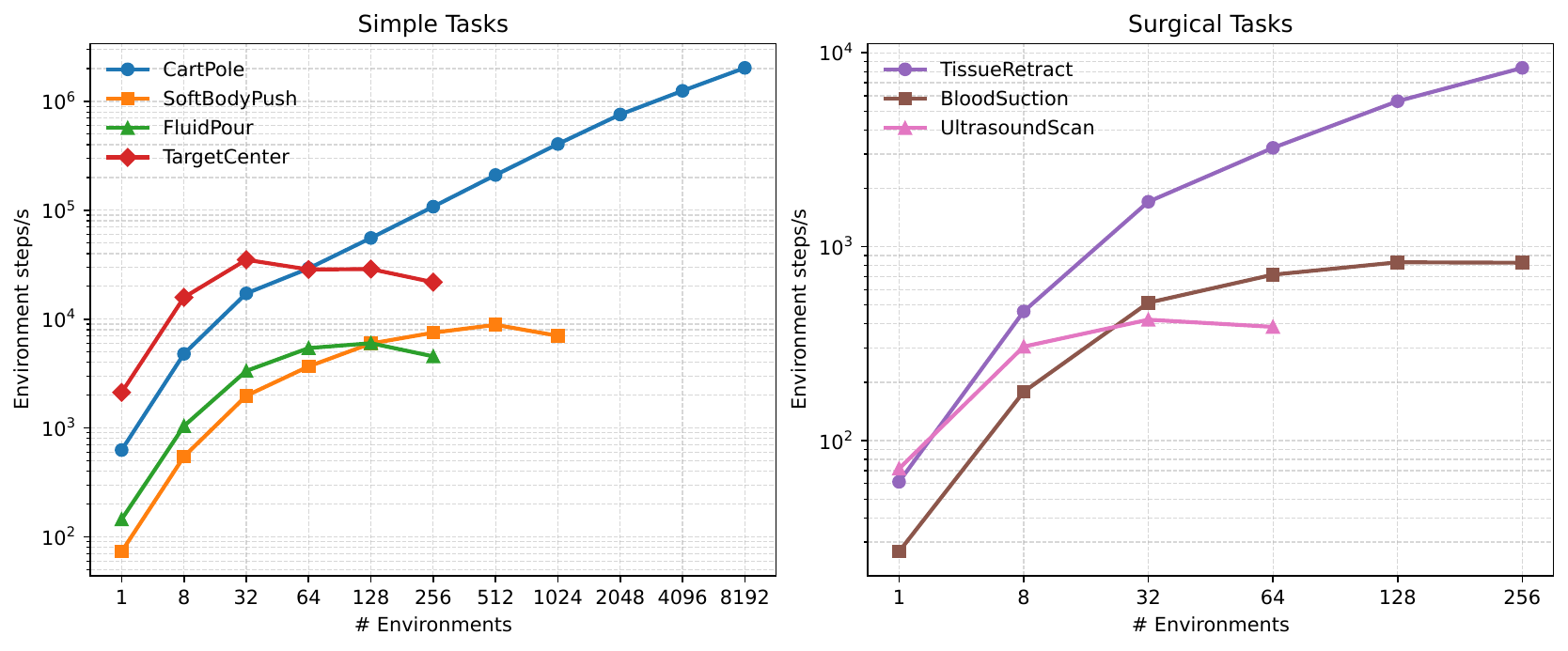}
}
\hfill
\subfloat[\label{fig:throughput_rl}]{
    \includegraphics[width=0.48\textwidth]{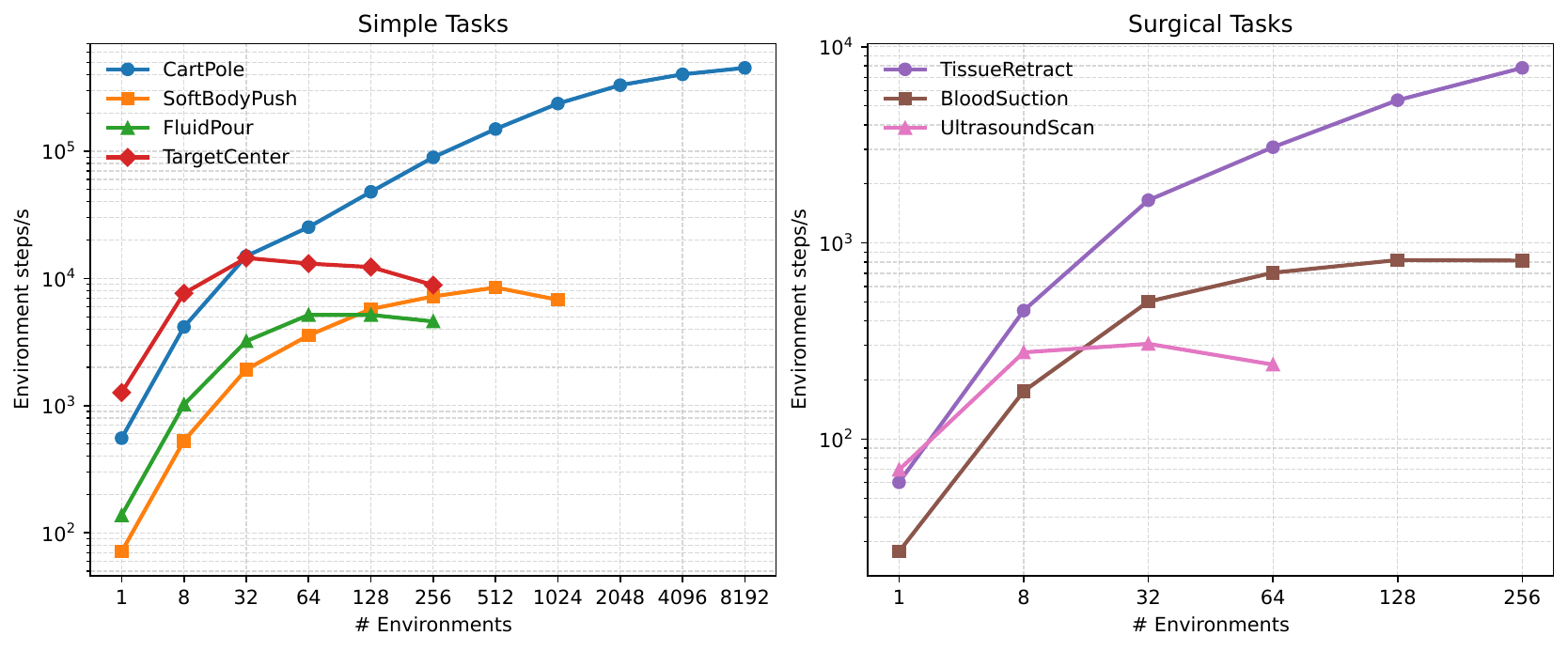}
}
\caption{Throughput scaling with the number of parallel environments on a single GPU: (a) pure simulation stepping throughput; and (b) end-to-end RL training throughput.}
\label{fig:throughput}
\end{figure*}

We evaluate throughput on the four simple RL environments, \texttt{CartPole}, \texttt{SoftBodyPush}, \texttt{FluidPour}, and \texttt{TargetCenter}, and the three surgical RL environments, \texttt{TissueRetract}, \texttt{BloodSuction}, and \texttt{UltrasoundScan}.
For each task, we report throughput as a function of the number of parallel environments on a single GPU.
Two settings are considered: (1) pure environment-stepping throughput, measured by repeatedly advancing each batched environment with zero actions and no policy updates; and (2) end-to-end RL throughput during training, which includes rollout collection and policy optimization.
Table~\ref{tab:throughput_setup} summarizes the benchmark setup, including hardware and other configurations, and the results are shown in Fig.~\ref{fig:throughput}.

\begin{table}[t]
\centering
\caption{Benchmark setup for throughput evaluation.}
\label{tab:throughput_setup}
\begin{tabular}{ll}
\toprule
Item & Value \\
\midrule
CPU & Intel Core i9-14900K \\
GPU & NVIDIA GeForce RTX 4090 \\
Host memory & 29 GiB \\
Operating system & EndeavourOS (Linux 7.1.8-arch1-3
) \\
Graphics backend & Vulkan \\
CUDA version & 13.3 \\
PyTorch version & 2.13.0 \\
Sim. warm-up & 64 env steps \\
Sim. measurement horizon & 256 env steps \\
PPO warm-up & 5 updates \\
PPO measurement horizon & 10 updates (128 steps/update) \\
RGB observation size & $64 \times 64$ \\
Ultrasound image size & $130 \times 160$ \\
Ultrasound probe scanlines & 96 \\
Ultrasound scanline length & 0.7 \\
Ultrasound scanline spacing & 0.006 \\
RL policy optimizer & Proximal policy optimization (PPO) \\
\bottomrule
\end{tabular}
\end{table}

As shown in the figure, throughput increases strongly with the number of parallel environments before reaching task-dependent saturation points.
The largest gains are seen in lighter rigid-body tasks.
For example, in pure stepping, the rigid-body \texttt{CartPole} task reaches 2.03M env-steps/s at 8192 environments, whereas \texttt{SoftBodyPush} reaches only 8.86k env-steps/s and saturates at around 512 environments.

In general, tasks achieve higher env-steps/s and scale to larger environment counts when their per-environment workload is low.
The saturation point is typically determined by the dominant per-environment bottleneck, whether in physics, rendering, sensing, or GPU memory usage.
For instance, \texttt{TargetCenter} saturates relatively early compared to \texttt{BloodSuction}, although both use RGB image observations.
One possible explanation is that \texttt{TargetCenter} already achieves relatively high throughput at small batch sizes.
As a result, image generation and image-space processing may become bottlenecks relatively early.
In contrast, \texttt{BloodSuction} has much lower overall throughput, so additional parallelism may continue to amortize parts of the fluid and particle computation, including per-particle suction logic, before performance begins to saturate.

End-to-end PPO throughput follows the same overall pattern.
Notably, PPO throughput remains close to pure stepping throughput for most of the heavier particle-based and surgical tasks, indicating that simulation dominates the runtime once physics and sensing become sufficiently expensive.

\section{Discussion}
\label{sec:discussion}

\subsection{Position of \projectref}
\label{sec:position_of_cressim_neo}

One might ask why another simulation engine is needed when hundreds of robotics simulators and platforms already exist\footnote{\href{https://github.com/knmcguire/best-of-robot-simulators}{\nolinkurl{https://github.com/knmcguire/best-of-robot-simulators}}}.
Even in the surgical domain, a number of recent studies have introduced simulation platforms, including ORBIT-Surgical \cite{yu2024OrbitSurgicalOpenSimulationFramework}, Surgical Gym \cite{schmidgall2024SurgicalGymHighperformance}, FF-SRL \cite{dallalba2024FFSRLHighPerformance}, LapGym \cite{scheikl2023LapGymOpenSource}, AMBF-RL \cite{varier2022AMBFRLRealtimeSimulation}, SurRoL \cite{xu2021SurRoLOpensourceReinforcement}, dVRL \cite{richter2020OpenSourcedReinforcementLearning}, many of which focus on machine-learning applications.
Instead of proposing another high-level platform built on an existing engine such as SOFA, Isaac Sim, or Unity, \projectrefbody\ is intended to provide a foundational simulation engine, on which surgical scenes and applications can be built.
At the same time, it is not intended to serve as a general-purpose replacement for existing engines, such as Isaac Sim.
Instead, it has a more focused position: a GPU-accelerated engine designed around the physical interactions and computational workflows particularly relevant to surgical robotics.

The engine deliberately emphasizes PBD-based simulation of rigid bodies, deformable tissues, fluids, strands, and manipulation-oriented interactions.
This differs from platforms that prioritize highly accurate rigid-body dynamics, differentiable simulation, or runtime GPU kernel generation through systems such as NVIDIA Warp or Taichi Lang.
Although these features are valuable for many general robotics applications, they are not central to \projectrefbody's design goals.
Therefore, the engine is better suited to manipulation-oriented surgical tasks than to general applications such as locomotion or navigation.

Conceptually, \projectrefbody\ can be viewed as a GPU-oriented extension of the simulation approach used by iMSTK and PositionBasedDynamics, augmented with batched rendering, multi-environment execution, and interfaces for robot learning.
Its strength is therefore not broad coverage of robotics domain, but a simple PBD framework that includes a wide range of capabilities for surgical simulation.

\subsection{Limitations}

\projectrefbody\ inherits limitations from both its underlying simulation method and its current architecture.
As discussed in Section~\ref{sec:position_of_cressim_neo}, the use of PBD makes the engine less suited to applications requiring accurate rigid-body contact dynamics.
In particular, rigid contacts and joints are less physically faithful than those provided by rigid-body engines (\eg MuJoCo), which can affect tasks such as rigid-object grasping or dynamical joint control.
In the current task setup, \texttt{CartPole} is included mainly as a benchmarking scene rather than as a practical target application.

At the implementation level, memory usage scales roughly linearly with the number of environments because many buffers are allocated according to the per-environment capacity multiplied by the environment count.
Additionally, topology-changing operations, such as adding particles or objects, still require buffer reallocation, scene rebuilding, or a re-upload of the simulation state.
In some applications, this can be mitigated by pre-allocating potentially required objects and keeping them disabled until needed, at the cost of additional memory consumption.

There are also known limitations that are planned to be addressed or mitigated as part of the release roadmap.
Custom compute passes can access a fixed set of exposed GPU buffers and render targets, instead of arbitrary simulator internals.
Some resources, including render-target textures, are currently exposed only for read access.
Cross-platform validation remains incomplete.
The actively supported graphics backends are Vulkan and D3D12.
CUDA interoperability is optional for the core engine but is currently required by ultrasound simulation.
Validation on AMD GPUs, MoltenVK/macOS, and systems without CUDA remains limited.
Finally, the rendering system uses a rasterized forward pipeline and does not support ray tracing.
Custom pipelines and graphics shaders are not yet supported.

Note that the RL examples are intended to demonstrate the learning interface and its computational performance, rather than RL policy performance. Task-specific learning and transfer are outside the scope of this work.

\section{Conclusion}
\label{sec:conclusion}

This paper presents \projectrefbody, a GPU simulation engine for surgical robotics and robot learning.
By combining PBD/XPBD with surgery-specific sensing and interactions, batched rendering, and GPU-resident data pipelines, the engine enables scalable simulation of surgical scenes.
The presented benchmarks and surgical task demonstrations show its potential for high-throughput learning and synthetic data generation in surgical robotics.

\bibliographystyle{IEEEtran}
\bibliography{BSTcontrol.bib,Paper-CRESSim-Neo.bib}



\end{document}